\documentclass[twoside,11pt]{article}

\usepackage{rawfonts}
\usepackage[cmex10]{amsmath}
\usepackage{algorithmic}
\usepackage{array}
\usepackage{stfloats}
\usepackage{url}
\usepackage[T1]{fontenc}
\usepackage[latin9]{inputenc}
\usepackage{color}
\usepackage{array}
\usepackage{float}
\usepackage{multirow}
\usepackage{amsmath}
\usepackage{amssymb}
\usepackage{graphicx}
\usepackage{longtable}
\usepackage[ruled]{algorithm2e}

\begin{document}


\title{Automatic Event Detection for Signal-based Surveillance}


\author{Jingxin Xu, Clinton Fookes, Sridha Sridharan \\
Speech, Audio, Image and Video Technology Lab\\
Queensland University of Technology
}

\maketitle

\begin{abstract}
Signal-based Surveillance systems such as Closed Circuits Televisions (CCTV) have been widely installed in public places. Those systems are normally used to find the events with security interest, and play a significant role in public safety. Though such systems are still heavily reliant on human labour to monitor the captured information, there have been a number of automatic techniques proposed to analysing the data. This article provides an overview of automatic surveillance event detection techniques . Despite it's popularity in research, it is still too challenging a problem to be realised in a real world deployment. The challenges come from not only the detection techniques such as signal processing and machine learning, but also the experimental design with factors such as data collection, evaluation protocols, and ground-truth annotation. Finally, this article propose that multi-disciplinary research is the path towards a solution to this problem.
\end{abstract}

\section{Introduction}

Since the invention of Closed Circuits Television (CCTV) in 1950s \cite{Bartlett1956}, signal-based surveillance systems have been widely installed in many places, such as airports, railway stations, bus stops, outdoor roadways, remote environments, industrial factories, health care facilities and schools. A signal-based surveillance system is a distributed sensor-based communications infrastructure. The sensors (e.g. cameras, microphones, etc.) are installed over a large region, and record information such as video, image and audio, which are sent over a communications channel to a receiver (monitor) that is usually located a significant physical distance from the sensors, enabling staff to monitor several areas and detect unusual events without physically being at the location. Nowadays cameras that capture natural vision information are the most widely used sensors. 

In the past several decades, the architecture and devices for surveillance have been largely improved. One main change is the wide-spread adoption of digital devices such as  IP (internet protocol) cameras, replacing their analogue counterparts \cite{Courtney2011}.   The utilization of digital devices allows computing techniques to play a more significant role, which includes the integration of surveillance  into a more comprehensive context aware information systems (such as ``Siveillance Vantage'' \footnote{http://www.buildingtechnologies.siemens.com/bt/global/en/security-solution/siveillance-vantage-command-control/pages/siveillance-vantage-command-control.aspx}).
At present such applications are heavily reliant on having human operators monitoring the incoming video feeds; however it is almost impossible to have sufficient human resources to monitor all the captured information (e.g. video feeds) continuously. Therefore, these surveillance systems are typically only used for post-event identification and forensic purposes after a crime has occurred. One such example is the investigation of the Boston Marathon Bombings \footnote{http://en.wikipedia.org/wiki/Boston\_Marathon\_bombings} in 2012.  CCTV footage was used to find the people who committed the crime after the disaster had happened, and even for this post events analysis, it is still fundamentally reliant on human labour. Ideally, CCTV systems would be monitored by a system that can detect events or interests as the occur, and fire alarms to alert human operators. Such a system would also need to be able to retrieve events from video in a rapid manner, to facilitate forensic analysis on surveillance video footage. 

This articles overviews techniques that are proposed for building an intelligent surveillance system which monitors public places of security interest, such as airports, railway stations, outdoor road ways, and so on, by automatically firing alarms upon the detection of unusual events. It is of interest to note that compared to event detection in other application domains such as sports analysis \cite{Rea2004,Sadlier2005}, broadcast television and movies \cite{Laptev2008}, event detection in a surveillance system has it's own properties due to the nature of surveillance systems. To facilitate further discussion, we first outline the four main characteristics of surveillance systems:

\begin{enumerate}
\item(\textbf{Property 1})A surveillance system is a communications system. Sensors such as cameras are typically located a long distance to the monitor that receives the captured signal.
\item
(\textbf{Property 2})
A surveillance system typically contains a database system, as the data is stored for a month or longer (depending on regulatory requirements) for forensic purposes.
\item (\textbf{Property 3})
A surveillance system is a continuous system,and typically operates 24 hours per day, 365 days a year.
\item (\textbf{Property 4}) A surveillance system is a critical system, with a responsibility to improve security and public safety.
\end{enumerate}
 
The properties of a surveillance system impacts the event detection techniques developed. For example, because of \textbf{Property 1}, \textbf{2} and \textbf{3}, data is typically compressed by a surveillance system. Due to \textbf{Property 3} and \textbf{4}, algorithms must be very robust. Besides the four main properties, individual applications have their own characteristics. In typical indoor CCTV deployments, cameras are most likely installed at fixed locations and are predominately static, leading to the popularity of   background subtraction techniques \cite{Lee2005,Piccardi2004,Zivkovic2004,DBM2012,JingxinPRL,Xu2012} in algorithm design, whereas in UAV (Unmanned Aerial Vehicle) surveillance, background subtraction becomes very challenging due to the motion of the UAV (i.e. camera) \cite{Talukder2004}.

The physical architecture, devices, and environment in which it is deployed can also all impact the algorithms used for event detection.  The algorithms themselves can be executed in several places. One option is to perform event detection at the sensor end, such that the functionality is embedded into the IP cameras \cite{Adam2008}. This offers only limited to the computational power, and thus algorithms have to be efficient with a low level of complexity. It is also possible to operate the algorithm at the receiver. Servers connected to the CCTV network can receive and process the feeds, performing the function of event detection \cite{Xu-IEEE-T-CSVT}, and allowing for a greater allocation of resources to the problem. In addition, it meets the requirements of modern information system architecture. The functions on such servers can be developed as the logical centric services on a Service Orientated Architecture (SOA). 

The type of environment also plays an important role. There are two main types of environment: indoor and outdoor; and each has its own unique challenges. For indoor surveillance, the installation of the cameras is restricted by the structure of the building. For example, it is often impossible to install a camera at a high location to support far field surveillance as the maximum height of the camera location is the ceiling height. Therefore, perspective distortion is typically severe (i.e. the same object has different size and velocity at different locations), which also leads to severe occlusions. For outdoor surveillance, a camera can be installed on top of a building and be positioned to look at the ground. The level of occlusion and amount perspective variation can be thus be reduced. However, in outdoor surveillance, there are usually complicated environmental and weather conditions. In some remote environments, the system installation is subject to the restriction of network facilities and power supply, which can influence the event detection techniques.   

In the context of the research literature, these different considerations are largely encapsulated by the choice of dataset. Specific datasets dictate specific applications, whether it be traffic surveillance or large indoor crowds. This choice determines not only the type of events being observed (which impacts on feature detection), but also the number and type of cameras, and often the type of learning model to be used due the type and amount of annotation provided. 

In this paper, we review current automatic surveillance event detection techniques and explore how the approaches that have been proposed are driven by the data that is available to the research community. We discuss the strengths and weaknesses of these techniques, and highlight several challenges that still remain open and prevent the wide-spread commercial adoption of techniques such as these. The remainder of this paper is organised as follows: we first provide an overview of the event detection process in Section \ref{sec:overview}, after which we outline the most common databases used in the literature in Section \ref{Datasets}. We then discuss the two main components of event detection systems, namely feature detection and the learning model, in Sections \ref{sec:event_representation} and \ref{sec:classifier} respectively. Finally, we conclude the paper in Section \ref{sec:Discussion}, and suggest avenues for future research to solve some of the outstanding problems in this field.

\section{Overview of the Event Detection Process}
\label{sec:overview}

\begin{figure}
\begin{center}
\includegraphics[width=12cm]{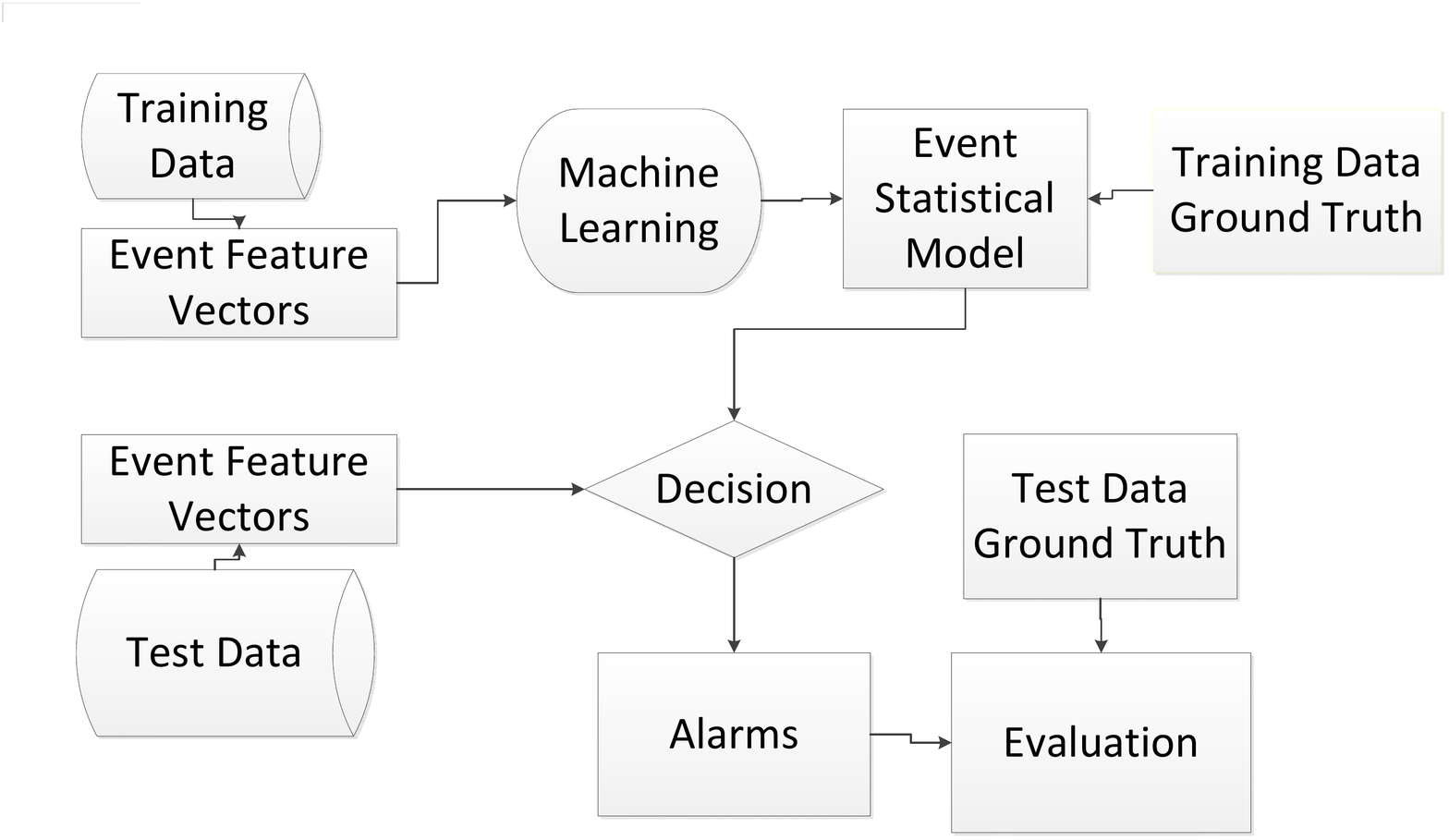}
\caption[Overview of Development Cycle]{This diagram illustrates the process of automatic surveillance event detection.\label{Fig:development-process}}
\end{center}
\end{figure}

Although the architecture and the environment play an important role in system design, the majority of event detection algorithms, irrespective of the events they are detecting, share a similar fundamental process. This section outlines this process, and it is illustrated in Figure \ref{Fig:development-process}. 

The events to be detected are governed by the data corpus used, and the data corpus is partitioned into two sets. One is used as the training dataset, and the other is used to test the system. The first step of event detection is extracting features to represent events. The purpose of this step is to find the features which are most effective in distinguishing the target unusual events from all others. After that, each event is represented as a feature set \footnote{Here we use the term ``feature set'' for its generality. The form of feature set is dependent on the machine learning techniques that are used in the following steps. Typically, a feature set can be represented in a vector form, which is termed a ``feature vector''. It can also be a set of correlated elements, which is termed a ``bag of words'' feature. When non-metric techniques such as decision trees are used, it is often referred to as set of ``attributes''. }. A feature set for an event is termed a ``sample'' or an ``instance''. Then the next step is to detect the samples associated with the target unusual events. This step relies on machine learning techniques. Fundamentally, there are two types of machine learning approaches: supervised learning, and unsupervised learning approaches.

In the application of event detection, if a supervised learning approach is used, ground truth annotation is required which labels the unusual events in the training dataset (i.e. a set of binary images which indicates the frames and regions in which in the unusual events occur). In such situations, one can train a statistical model that captures the characteristics of the unusual events. Given a sample of the test dataset and the trained model, the distance between the sample and the trained model is computed, and based on a threshold, a decision is made. If there are also annotations for the usual events in the training dataset, one can train another statistical model for the usual events. Given a sample of the test dataset, the system can compare the test sample to both the statistical models. Then, if the sample exhibits a greater similarity to the unusual event model, the event is classified as an unusual event.

On the other hand, unsupervised learning approaches would be used if the training dataset contained only usual events. The learning process trains a statistical model from the training dataset where only the usual events are observed. Given a sample of the test data, the similarity between this sample and the trained statistical model is computed for usual events. If the similarity is lower than a threshold, this sample is deemed to be an unusual event.

Both supervised and unsupervised approaches require ground truth for the test dataset in the algorithm design cycle. The alarms raised by the systems are  compared to the ground truth of the test dataset for performance evaluation. Here we use the term ``metric'' to define the distance between a sample and a trained model, which is a measure of the difference between an individual sample from the test dataset to the distribution of the training samples in the feature space. For generative models, the metric is a probability. In most situations, a feature set is represented as a high dimensional vector, and there are correlation dependencies among the elements in the set. There are also some machine learning models which receive a set of independent elements, notably the ``bag of words'' models. It is also popular to use rule based classifiers for event detection, such as decision trees, learning association rules and grammars. For these classifiers, features are extracted as a set of attributes.

Besides supervised and unsupervised learning, recently there are methodologies \cite{Hospedales2011} that use so-called ``weakly supervised learning'' techniques. The ``weakly supervised learning'' techniques are in fact a special type of supervised learning. It differs from traditional supervised learning methods as the annotations are marked at a coarse level. For instance, only temporal labels (when the event happens) are available in the training dataset, but no information for where the event is happening.

\subsection{Popular Research Datasets and Evaluation Methodologies}
\label{Datasets}

A large number of datasets have been published within this field, covering a variety of situations including mass panic \cite{WEBDataset,UMNdataset}, traffic surveillance \cite{VIRATVideoDataset,IDIAPJunctionDataset}, and pedestrian/person surveillance \cite{PETS2009,Zhou2012,Train_station}. Many of the most popular datasets are summarised in Table \ref{tab:datasets}, and Figure \ref{fig:dataset} shows sample images from the datasets.

\begin{longtable}{|p{2cm}|p{3cm}|p{1.5cm}|p{3cm}|p{2cm}|}
\hline
Name & Length & Resolution & Event Types & Ground Truth \\
\hline
UCSD Anomaly Dataset (Peds 1) \cite{Mahadevan2010} & 34 Training Examples, 36 Testing Examples & $158\times238$ & Pedestrian Anomalies (i.e. cyclist, skateboarder) & Yes (Frame Level for all; a subset  with Location Level) \\
\hline
UCSD Anomaly Dataset (Peds 2) \cite{Mahadevan2010} & 16 Training Examples, 12 Testing Examples & $360\times240$ & Pedestrian Anomalies (i.e. cyclist, skateboarder) & Yes (Frame Level for all; a subset with Location Level) \\
\hline
PETS 2009 \cite{PETS2009} & 11 training time sequences, 18 test time sequences (9 for event detection), 8 cameras are used  & $768 \times 576$ or $720 \times 576$ & Various group level events (acted) & Partial (no ground truth provided with the original release, though various authors have released their own annotations)  \\
\hline
PETS 2014 \cite{PETS2014Dataset} & 24 scenarios with 22 scenarios with acted abnormal behaviours &$1280 \times 960$ & abnormal and criminal events in a parking area & Yes (video level) \\
\hline
UMN Dataset \cite{UMNdataset} & Approx 8000 frames across 11 sequences & $320 \times 240$& Mass Panic (acted) & Partial (no ground truth provided with the original release, though various authors have released their own annotations)  \\
\hline
WEB Dataset \cite{WEBDataset} & 20 short sequences (12 training, 8 testing) & various & Events in heavy crowds (fighting, mass panic) & Yes (Sequence Level) \\
\hline
MIT Traffic Dataest  \cite{MIT_Traffic_dataset} & 90 mins & $720 \times 480$ & Traffic events at an intersection & Partial (no ground truth provided with the original release, though various authors have released their own annotations) \\
\hline
QMUL Traffic Dataset\cite{QMULTraffic}& 3 separate approximately 1 hour scenes. Segmented into $150$ $X$ second clips each & $360 \times 288$ & Traffic events at two intersections and one roundabout & Yes (Clip Level)\\
\hline
TRECVid SED Dataset \cite{2012trecvidover}& 144 Hours & $720 \times 576$ & Pedestrian events at an airport (people meeting, people splitting, embrace, using a mobile phone, etc.) & Yes (Frame Level), though only the first 100 hours is publicly available \\
\hline
Subway Dataset \cite{Adam2008}& Two video files, one is for the entrance and the other is for the exit & $512 \times 384$ & wrong direction (e.g. entering a subway from the exit end); no payment  & Yes (Frame Level). Some researchers released new ground-truths for the events in their applications.\\
\hline
VIRAT Ground Dataset \cite{VIRATVideoDataset,Oh2011a}& 11 hours video files for 16 outdoor scenes & $1280 \times 720$ & single agent events (e.g. running); multiple agent interaction (human-vehicle events) & Yes, fine annotation \\
\hline
IDIAP Junction Dataset \cite{IDIAPJunctionDataset}& 44 minutes &$360 \times 288$ & Abnormal Events in Traffic Surveillance& Yes, the time of the abnormal events is provided. \\
\hline
Train Station Dataset \cite{Zhou2012,Train_station} & 30 minutes video recording from New York Grand Central Train Station& $720 \times 480$ & The target events are not defined & No \\
\hline
\caption{Summary of commonly used datasets.}
\label{tab:datasets}
\end{longtable}






Although numerous datasets are available, as can be seen from Table \ref{tab:datasets} the majority are small in size. The small size of many datasets make evaluation difficult, and leads to unreliable results. Only the TRECVid dataset can be viewed as a dataset that could produce statistically significant results. However, even the TRECVid dataset has significant limitations. For example, the events identified  in the TRECVid datasets such as people meeting and people splitting are normal behaviours in any airports. There is a lack of events defined in the dataset that are of real security interest. 

The reason for this lack of large databases is simple: the collection and annotate of a large, realistic surveillance event database is very challenging. There are several reasons for this. For instance often the most interesting events in a real-world context (i.e. drunk drivers in traffic surveillance) are difficult, if not impossible to obtain in a controlled manner, requiring thousands of hours of footage to be collected to capture even a few instances. Having such an event acted out is not possible either, due to the security and safety implications.

Another complication is the ground truth annotation. The quality and type of ground truth annotation available varies, and it it not uncommon for some databases to have multiple versions of ground truth annotation. For instance, although the MIT Traffic Dataset \cite{Wang2009PAMI,MIT_Traffic_dataset} was not originally provided with ground truth, ground truth has been annotated by \cite{Hospedales2011} and later \cite{JingxinPRL}; although for different purposes with \cite{Hospedales2011} providing annotation for anomalies while \cite{JingxinPRL}  annotates all instances of a set of four events (traffic turning and pedestrians crossing the road).

Ground truth is also inherently difficulty to annotate due to the ambiguity in the events. Ideally, all events would be annotated with time and location information (i.e. bounding boxes or even binary masks for every event, every frame), however, aside from this being clearly impracticable for any dataset of substantial size, correctly identifying when an event starts and ends is highly ambiguous. As such annotation rarely includes location information, and is performed at the frame, or even clip (i.e. short sequence) level. The lack of fine detail (i.e. locations) means approaches that use supervised learning are difficult to develop, with the exception of scene wide events such as those depicted in the UNM \cite{UMNdataset} and Web \cite{WEBDataset} datasets. Variation in the ground truth also makes it difficult to fairly compare the performance of different approaches across different databases.

The ground truth annotation is not only important for the training of models, but also for their evaluation. The majority of researchers compute receiver operating characteristic (ROC) curves or detection error trade-off (DET) curves based on the true and false alarm rates of the frames or video clips, rather than for the individual events. This seems somewhat counter-intuitive (particularly at the frame level), given the difficulties in annotating the start and end of events. For instance, suppose we have two algorithms and are evaluating on a dataset with two events, each of which contains two frames. If the first algorithm detects the two frames in the first event and misses the two frames of the second, and the second algorithm detects one frame from each event, both the achieve the same true detection rate if a frame level ground-truth is used. However, from a practical point of view, the second algorithm has performed better, as it detected at least part of both events. This highlights a vulnerability is existing widely used evaluation methodology.  Thus even if ground truth is annotated at the frame level, it is more sensible to evaluate the systems at the event level. 

The TRECVid SED competition  \cite{2012trecvidover,2011trecvidover} proposed a concept of ``event alignment'' to address this problem. There are four steps for the TRECVid evaluation approach:
\begin{enumerate}
\item Segmentation mapping using the Hungarian algorithm to assign detected events to ground truth events;
\item Segmentation scoring, including correct detection rate, missed detection rate, and false alarm rate;
\item Computation of the Normalised Detection Cost Rate (NDCR): NDCR=0 indicates a perfect system, and NDCR=1.0 indicates a system equivalent to no output; and 
\item Visualisation using Detection Error Trade-offs (DET) Curves.
\end{enumerate}

The evaluation methodology proposed in the TRECVid SED task solves the problem to some extent, but it is not a thorough solution, as there are problems if two target events happen at the same time. The protocol so far hasn't considered these exceptional situations.

Besides the dataset that purely based on visual data, there are datasets with acoustic features. The BOSS dataset contains visual-audio surveillance footages of 15 sequences captured by 9 cameras and 8 microphones. Activities such as cell phone theft, panic, etc. are defined. There is also a sequence with camera mul-functions features in this dataset. There are also datasets with pure acoustic information. The MIVIA \cite{MIVIADataset} contains 20 hours audio training dataset and about 9 hours test dataset. The abnormal acoustic events include screams, gunshot, and glass-breaking. The IEEE AASP Challenge Dataset \cite{IEEEAcousticDataset} is also a very popular one for acoustic scene classification and event detection. For event detection part, the dataset consist of two partitions. The first partition is recorded from real-life scenes with non-overlapping events, such as ``door knock, door slam, speech, laughter, keyboard clicks, objects hitting table, keys clinging, phone ringing, turning page, cough, printer, short alert-beeping, clearing throat, mouse click, drawer, and switches.'' The second partition is synthesized using the first partition, and different events are overlapped in time. Recently, the PETS 2015 \cite{PETS2015Dataset} dataset has been released. The PETS 2015 \cite{PETS2015Dataset} consists of the PETS 2014 dataset and a new dataset of video files captured by three visual IP cameras and four thermal cameras near a nuclear plant. Target activities in this dataset include vehicle driving, boat driving, throwing and fetching bags, people meeting and splitting.

\section{Surveillance Event Detection Techniques}
\label{liter_activity_perception}

This section reviews the techniques for surveillance event detection. From Figure \ref{Fig:development-process}, the design of an event detection algorithm consists of two fundamental problems:
\begin{enumerate}
\item the design of feature descriptor to represent the event; and
\item the design of the classifier to detect/recognize the event.
\end{enumerate}

We address these two problems in subsections \ref{sec:event_representation} and \ref{sec:classifier} for feature extraction and classifier training respectively. It is also of interest to note that, not all event detection methods utilise statistical models. For example, it is possible to simply set a threshold on sound volume to detect a gunshot event; and speed related events (i.e. a speeding vehicles) can be detected by applying a threshold to a measured velocity. 

\subsection{Feature Extraction and Event Representation}
\label{sec:event_representation}

The design of a feature vector to represent the events is highly dependent on the target application. For instance, motion features which capture objects' velocities are useful to detect a person running event; whereas the detection of a person wearing a hat has to rely on appearance features. The fundamental aim in this step is to extract features that can separate the usual and unusual activities from one another. Furthermore, the feature descriptor should be robust to variations of the same event. Once the events are represented by a certain feature set, there may be a pre-processing step to transform the data into another feature set (e.g. dimension reduction).

Feature extraction methods can broadly be categorised into the following groups:
\begin{enumerate}
\item local features; 
\item feature descriptors from trajectories; 
\item feature descriptors for collective behaviours; 
\item feature descriptors from compressed domain; and
\item acoustic feature. 
These five types of features are discussed in the following five subsections.

\end{enumerate}

Then in the following sections, each type of feature will be reviewed in detail. 

\subsubsection{Event Representation by Local Features}

This section reviews methods that directly extract features from a local region, which we term ``local feature''.

Early publications for activity perception are typically for non-crowded scenes, and objects' trajectories extracted from a tracking algorithm are often used to represent the events \cite{Liao2006,Johnson1996609}. The trajectories have the advantage of capturing the objects' long term motion, thus these algorithms are successful when object tracking is robust. However, real world surveillance scenes are often crowded, and due to the occlusions and clustering, robust object tracking is challenging. Thus recently, research in this field has focused on adopting features beyond tracking \cite{Adam2008,Andrade2006,Hospedales2011,Mahadevan2010,Ryan2011,wangabnormal,Wang2007,XiangG06}. 

One of the early works for event detection beyond tracking approaches is \cite{XiangG06}, where a descriptor termed the ``pixel change history'' (PCH) is proposed. A pixel's PCH is defined as 

\begin{equation}
P_{\zeta,\tau}=\begin{cases}
min(P_{\zeta,\tau}(x,y,t-1)+\frac{255}{\zeta},255) & \quad(D(x,y,t)=1)\\
max(P_{\zeta,\tau}(x,y,t-1)-\frac{255}{\tau},0) & otherwise
\end{cases}\label{eq:PCH}
\end{equation}

where $P_{\zeta,\tau}(x,y,t)$ is the PCH for a pixel at location
$(x,y)$; and $D(x,y,t)$ is the binary mark for the foreground image.
The parameters $\zeta$ and $\tau$ are the accumulator factor and
decay factor respectively. When $D(x,y,t)=1$, the PCH will increase
gradually, at a rate controlled by the accumulator factor $\zeta$; and when $D(x,y,t)=0$, the PCH will decrease gradually at a rate controlled by the decay factor $\tau$. In this way, the PCH records temporal changes caused by moving pixels. Based on the extraction of the PCH, salient pixel groups are identified and a 7-dimensional feature vector is defined (the feature vector is formed by some statistical properties of PCH and the pixel location). Then a Gaussian Mixture Model (GMM) is trained to cluster the salient pixels into different events in an unsupervised manner. However, despite this method is beyond tracking, the evaluation data shown in the paper is in a sparse scene. There is still lack of evidence for its performance on crowded scenes where object tracking is not reliable. 

Methods based on local features often rely on the computation of optical flow\footnote{Optical flow estimation is a computer vision technique that computes pixel's motion vector between two successive frames.} \cite{Adam2008,Andrade2006,Hospedales2011,li2011learning,Wang2007,Wang2009PAMI}. 
A popular descriptor \cite{Wang2009PAMI,Hospedales2011}
encodes the moving pixel's location and velocity into a discrete
codebook, where the velocity is computed using optical flow estimation. Each element from the codebook is called a codeword. 
Typically, this feature descriptor is used along with probabilistic topic models \cite{Blei2010,Blei2003}. The probabilistic topic models learn statistical properties of the code words in the training datasets and extract distributions of the codewords as the topics. In the application of activity perception, each topic represents an event.  In Wang et. al \cite{Wang2009PAMI}, moving pixels are first detected by a subtraction operation over every two successive frames. Each moving pixel will result in an entry in the histogram of the code words. However, the number of code words for the training dataset is often very large due to the large number of moving pixels, making it impractical to be trained when Markov Chain Monte Carlo (MCMC) sampling is used as the inference method\footnote{For certain kinds of topic models, MCMC sampling is a by default choice for inference.}. Thus in \cite{hospedales2009markov}, a descriptor derived from this feature is proposed which encodes the mean of the moving pixel's location and direction over a small patch. In this situation, the number of total code words is reduced and this feature is applicable to a wider range of applications. Adam et. al \cite{Adam2008} proposed a fast algorithm for unusual event detection using local monitors. The term of ``local monitors'' means a set of local points. Each point was allocated a buffer which stores the information of optical flow in the recent duration. Statistical properties of the stored optical flow in each monitor are used for the detection of low likelihood patterns as the unusual events.  One key limitation of optical flow is the so called ``aperture problem'', as the estimation of optical flow is vulnerable at textureless regions. The second limitation is that the motion features are extracted based on every two successive frames, which does not capture events that have a longer duration.
Despite of its limitations, the methods based on optical flow typically have high performance on detecting events defined by moving directions, such as a traffic turn \cite{Wang2009PAMI}, or wrong direction movement\cite{Adam2008}. 

The feature reviewed so far are based on pixel level.
An alternative to pixel level features is to extract features from patches. Spatio-temporal patches \cite{Mahadevan2010,Ryan2011,yang2011} better capture temporal information about the event, while some approaches \cite{Mahadevan2010,Kellokumpu08humanactivity,Yunqian2009,Jingxin_DICTA} 
can also capture appearance features. In these approaches, the video is cut into a regular space-time grid. Each grid square is called a spatial-temporal patch. In \cite{Mahadevan2010}, the events are represented as mixture of dynamic textures from the spatio-temporal patches. Dynamic textures are sequences of images of movement that exhibit spatio-temporal stationary properties, which is modelled by  auto-regressive moving average processes (ARMA) \cite{Doretto2003}.  Chan and Vasconcelos
\cite{Chan2008} proposes the Mixture of Dynamic Textures (MDT) to allow the modelling of multiple co-occurring dynamic textures (e.g. fire together with smoke), and this technique is used in Mahadevan et. al \cite{Mahadevan2010} for detecting anomalous events in crowded scenes by considering
both temporal abnormalities and spatial abnormalities. However, the computational complexity of this approach \cite{Mahadevan2010} is high, limiting it's use in real world conditions. Ryan et. al \cite{Ryan2011} extended a traditional image texture technique into the optical flow field, and defined a new descriptor termed ``textures of optical flow'', with the motivation that the texture of optical flow for vehicle motion is smoother compared to those generated by pedestrians. In \cite{yang2011} a multi-level histogram of optical flows is proposed in the spatial temporal patch framework and sparse coding is used as the classifier. The typical limitation of spatial-temporal patches is that the cutting of patches may separate an event into multiple patches which causes noise and inaccuracies. Meanwhile, the size of the patch is hard to determine, as a uniformly sized patches will encounter problems in the presence of perspective distortion.  An alternative to using regular patches is to use a sliding window approach. This approach is adopted in \cite{Ke2010}, where shape feature is used. This method is shown to be effective to detect events such as hand shaking. However, the sliding window matching is time consuming, especially for feature descriptors with high computational complexity (e.g. histogram of gradient orientation, optical flow, etc.). The sliding window matching approach is also used in  \cite{Yuan2009}, where it is termed a ``sub-volume''. In \cite{Yuan2009,Tran2014}, an more efficient searching algorithm is proposed, which is similar to track a three dimensional feature patch over time and space. It is of interest to note that, compared to regular patches, the sliding window approach potentially can introduce more noise from the background. In regular patches, the background pattens in a patch over time are similar to each other. In sliding window approaches, the background variation is expected to be much more serious.

With the popularity of key point detectors and descriptors such as SIFT \cite{SIFT} and SURF \cite{SURF} in image retrieval, spatio-temporal analogues such as 3D SIFT \cite{Scovanner2007} have been proposed. Key point detection aims to find the locations that are best for feature extraction, and subsequently extract features at such locations. The feature that is used to find such a location is also used as the feature for event detection. In \cite{Chen2009}, a type of SIFT extension called ``MoSIFT'' is proposed, by adding the temporal motion information. This feature is shown to be useful to detect local events defined by the TRECVid SED competition. In  \cite{Cao2010,Tian2012}, key point detectors such as Space-Time Interest Points (STIP) and its extensions are used as the feature descriptor. By using the universal background Gaussian Mixture Model, these method can detect an event in a scene that is different from the training dataset. This technique is termed ``cross-dataset action detection''. Zhao et al. \cite{Zhao2011} proposed an approach of online unusual event detection using sparse coding. In this approach, space-time interest points are first detected to find the regions of interest, and then HoG (histogram of Gradient) and HoF (Histogram of Optical Flow) are extracted as the features for the detection algorithm.  From theoretical point of view, almost all feature detector aims to find the ``corner'' (saliency) in the feature space. However, there is lack of evidence to support whether the abnormal events have to be in such feature corner location. Therefore, the miss detection of the feature location leads to the miss detection of the event at that location.

More recently with the popularity of deep learning networks, there have been attempts to use unsupervised feature learning for event detection in crowded surveillance video \cite{Ji2010}. The approach of \cite{Ji2010} can be viewed as the combination of a key point detector and a patch-based feature. The input to the deep neural network is the raw pixel intensities; the output of the first layer can be viewed as a set of key points. The features are learned in earlier layers and the classifier is contained in the last layer. The benefits of this approach include the avoidance of user defined features, which allows the learning algorithm to select optimum features for the events. However, this approach requires the use of a very large dataset. The dataset used in \cite{Ji2010} is the TRECVid SED dataset containing 100 hours of video. However, even though this is a large amount of data, it is still not large enough for a fully ``unsupervised feature learning'' as the user still needs to initialize the parameters with some state-of-the-art handcrafted features.  In addition, the processing of the dataset in this method is impractical for real world applications due to its computational cost. 

In \cite{cheng2015video,wang2015video}, key point detectors are used as the preliminary step to extract features that can model mutual interactions. In \cite{cheng2015video}, space-time interest point (STIP) detectors are first used to detect the key points and generate the low level codebook which is used for local anomaly detection. Then based on the geometric constraint of the STIPs, a higher level codebook is generated, followed by a Gaussian process regression for global anomaly detection (interaction level). In \cite{wang2015video}, the STIPs are used as the baseline features, and SIFT key point detectors are used to obtain the appearance and interaction based context features. Then a novel context model based on deep architecture is proposed to fusion the different features for event detection.

Extracting features based on a key point detector can capture the locations that are most suitable for feature extraction. Furthermore, a key point descriptor is typically able to capture rich information (i.e. motion, appearance) which is robust to view point and scale variations. However, the missed detection of the key points will result in the missed detection of events; in the application of unusual event detection, missed detection is more costly compared to false alarms. 

Hasan et. al \cite{Hasan2015} proposed using object tracking by detection techniques to find the locations of interest, and then space-time interest points (STIP) features are extracted as the input to a deep learning architecture. The output of the deep learning feature is further used as the feature for the classifier to detect events of interest. Limitation of this approach is that the performance is highly sensitive to the object tracking performance. 

In summary, local features are the most widely class of features for event detection in crowded scenes. This is because traditional features for event detection in non-crowded scenes, such as the objects' trajectories, are difficult to extract in crowded environments (i.e. due to the failure of object tracking). However, local features are not suitable to model events which happen over a large area as individual features view too small a part of the event.


\subsubsection{Trajectory Based Event Representation}

Trajectories that are obtained from an object tracking algorithm are often used as a feature to represent events \cite{Stauffer2000,Sefidgar201516,Ji2010}. In \cite{Stauffer2000}, a real time tracking algorithm that is based on background subtraction is proposed, and the trajectories obtained are used to identify activities in traffic surveillance footages. However, tracking algorithms based on background subtraction to detect the target object usually have poor performance in crowded scenes, which limits the approach in \cite{Stauffer2000} to be applied in more general applications. A similar approach is presented in\cite{Hu2006}. However, the most interesting point in \cite{Hu2006} is in fact the use of radio control toys to conduct experiments. It is well known that abnormal events in traffic surveillance are normally hard to record in the real world. The use of radio control toys is a rational approach for experimental design. 

Due to the development of object detection such as the histogram of gradient orientations (HOG), there has been algorithms that can track all individuals in very crowded scenes such as those in PETS 2007 dataset in real time \cite{benfold2011stable}. As a result, recently there are publications of using trajectories from object tracking for event detection, such as \cite{Sefidgar201516} and \cite{Ji2010}.However, it is important to point out that, object tracking based on object detection has its limitation in the application of event detection. The preliminary requirement of such approach is to train an object detector in supervised learning. The missed detection and false alarm in object detection have strong influence on the following tracking performance. Furthermore, in a complicated scene, there are normally a number of different types of objects, such as pedestrian, bicycle, bus, car, animals, cart and so on. To be able to model activities among those objects, detectors for all the types of objects should be trained, which is impractical in many real world applications.

Though object tracking is not robust in crowded scenes, the idea of capturing long term motion features is desirable. As a result, the trajectory-based features have attracted considerable interest and approaches have been proposed to extract trajectory feature from crowded scenes including using particle trajectories \cite{wu2010chaotic}, or by reconstructing the corrupted trajectories \cite{zhourandom}. 

Trajectories have several unique benefits for modelling events in video, which are  difficult to replicate with other feature extraction approaches. A major advantage of trajectories is that they can model longer duration motions compared to other features such as optical flow. In addition, trajectories model motions in a larger space compared to local features. For instance, in the spatial-temporal patch approaches, the pattern captured is always within the space of a single spatial-temporal patch, whereas trajectories can move across a set of different patches. 

There has been a long history of the use of trajectories as a feature for activity recognition. Early investigations for event detection were typically for non-crowded scenes, and used trajectories generated by object tracking \cite{Liao2006,Johnson1996609}. The popularity of approaches relying on object tracking has declined in recent years, as research has focused on crowded scenes where object tracking is less reliable. However, this approach is still pursued in situations where there is minimal crowding \cite{Wang2011IJCV}. An alternate approach is to try to reconstruct corrupted trajectories to overcome the problems caused by tracking failures. Zhou et. al \cite{zhourandom} applies Markov random fields and spanning trees to link the incomplete trajectories terminated by occlusions in crowded scenes. However, it should be pointed out that the surveillance scene in which this algorithm has been evaluated is quite unique. The camera was installed in the ceiling of a tall building (New York Grand Central Station). As a result, the level of perspective distortion is very low, which minimises occlusions. Another dispointed point for this approach is that, the dataset that used in the evaluation does not contain abnormal events with security interest. There are no qualitative evaluations in the article.

Recently, there are also methods proposed using point trajectories \cite{ali2007lagrangian,Takahashi2010}. In \cite{ali2007lagrangian}, the trajectories are approximated by optical flow and this approach also links to the physical model-based representation discussed in the next section. In \cite{Takahashi2010}, the SIFT detector is first applied to detect the SIFT key points. Then these key points are tracked using the KLT tracker \cite{KLT}. The feature vector is designed to encapsulate the key point trajectories; the method is evaluated using TRECVid dataset and it is said to achieve state-of-the-art performance. The key limitation of this approach is that, the SIFT detector is a feature detector in two-dimensional image domain. It is possible to have some SIFT points in background regions, potentially causing false alarms; it is also possible to have moving object without SIFT points detected. This will lead to missed detection.    Recently, Xu et al. \cite{Xu2012,JingxinPRL}
 proposed using a Fourier based trajectory descriptor for event detection. This include applications for abnormal event detection \cite{Xu2012} and known event detection\cite{JingxinPRL}.  In these approaches, point trajectories are constructed. Each trajectory is a two-dimensional signal, and based on which, a Fourier transform is performed. The Fourier coefficients of low frequency components are used to form a feature vector, followed by a K-means clustering to discretize the feature into a ``bag of words''. Then various machine learning methods are investigated for event detection based on this feature descriptor. The use of Fourier transform effectively capture the shape of the trajectory, as the low frequency Fourier component preserve the fundamental shape of the flow. Therefore, this feature is competent to detect events such as a traffic turn. However, the use of Fourier transform only captures the global properties of the trajectories with the region based properties lost.  Xu et al. \cite{Xu-IEEE-T-CSVT} proposed a feature that work formed by the angles and distances between pairs of trajectories, with the aim of detecting interactions between people (in particular, the `people meet', `people split' and `embrace' events in TRECVid). State-of-the-art performance was achieved for these events, however performance was very poor for the other four events in the TRECVid dataset, highlighting the limitations of the feature.

Though the extraction of point trajectories is more tolerant to noise in crowded environments, point trajectories are unable to distinguish between activities performed by different objects in a clear manner. Furthermore, the existence of perspective distortion leads to imbalanced numbers of trajectories for activities occurring in the near and far field, which will potentially cause inaccuracies in detection. This issue has been raised in \cite{Xu2012}.

It is interesting to note that event detection is not always a subsequence process of a tracking algorithm. In practice, it can also be used the initiation of an object tracking algorithm. This is partly because tracking all individuals in a complicated scene are often challenging, while tracking one to two individuals in the same scene is usually much easier. In \cite{huang2015boost}, the objects that of the abnormal behaviours are first detected by using sparse coding on the multi-scale histogram of optical flows, and then the locations of the detection are used as the windows for an object tracking algorithm.

\subsubsection{Feature Extraction for Collective Behaviours}

Under specific environments, pedestrian movements can be modelled using a physical based model \cite{ali2007lagrangian,mehran2010streakline,Mehran2009,Kaltsa2015}. The suitability of these approaches depends on how  well the model matches what is observed in the physical world.

This section reviews approaches which model the crowd movement using physical models. The motivation for the early work \cite{ali2007lagrangian} in this direction is that, in surveillance scenes with high crowd densities, the movements of pedestrians are similar to the motion of water currents, allowing physical models from fluid dynamics to be used. Approaches in this area include modelling a pedestrian in highly density far field surveillance scene as a fluid particle in the Lagrangian dynamic process \cite{ali2007lagrangian,wu2010chaotic}, or modelling a group's motion as streamline or streakline \cite{mehran2010streakline}. Solmaz et al. \cite{Solmaz2012} models the crowds as a dynamic systems to identify five classical crowd behaviours. 

The physical models applied in this field potentially include particle to particle interactions. The underlying theories in these physical models relate to the force and motion of those particles. Correspondingly, in extremely crowded scenes where these techniques are used, the models deal with the force and motion relationships among pedestrians. It is assumed that a single pedestrian's activity is influenced by others in high crowd density environments. The motion patterns that correspond to the activities from a group's perspective are called the crowd's collective behaviours. There is evidence in other research domains such as complex systems \cite{helbing1995social} that an individual's motion is influenced by the group's motion, and a single pedestrian is likely to move with the flow of a group with a level of variation which reflects the person's own destination, and the intention of keeping a comfortable distance from each other. These mutual influences can be modelled by approaches such as the social force model \cite{helbing1995social}. Recently the social force model was applied to model crowd activities \cite{Mehran2009}. In this method, the video is partitioned into clips with a small number of frames. A set of particles are located in the scene, and then short duration trajectories are constructed by approximating the optical flow field. The fourth order Range Kutta algorithm is used in the approximation step. The social interaction forces are computed based on the social force model. The forces are then quantized into discrete code words. The unusual events are detected using Latent Dirichlet Allocation \cite{Blei2003}, as the low likely video clips are detected as those containing abnormal events.  Recently, Xu et. al.  \cite{DBM2012} proposed to use the distributed behaviour model, and outperforms the social force model in the situation of camera movements. More recently, Zhang et. al extends the social force model to social attribute force model \cite{Zhang2015} by introducing social disorder and congestion attributes to characterize the social interactions. However, since the focus of the approach in \cite{Zhang2015} is the interaction level events, the performance for this approach to detect local abnormal events that are defined in the UCSD dataset is lower than some state of the art method.

There are similar approaches to model the crowd movements as other physical processes, including \cite{cui2011abnormal} which defines the concept of interaction potentials, and \cite{Kaltsa2015} which proposed a feature called ``Histogram of Oriented Swarms'' to capture the crowd dynamic and detect abnormal behaviours. 

Besides using physical-based models, recently there have been some methods proposing pure mathematical models to capture the features of collective behaviours, including using Lie Algebra and geometric flows \cite{lin2010modeling,lin2009learning}. These approaches require the surveillance scene to be captured in the far field. Meanwhile, the success of these techniques highly depends on level the crowd density. Higher crowd densities generally fit this model better. For instance, datasets used in \cite{lin2010modeling,lin2009learning,cui2011abnormal,ali2007lagrangian,wu2010chaotic,Solmaz2012,Mehran2009} are all very crowded. The most popular application for this direction is the detection of rapid escape event. However, as real world surveillance scenes usually contain a mixture of different levels of crowd density, the requirement of a high crowd density limits the effectiveness of these approaches to specific datasets or problems.

\subsubsection{Extracting Features from video compressed domain}

A recent article \cite{Huang2014} indicates that  surveillance video data is the biggest of the `big data' problems. For forensic applications, faster than real-time performance is required \cite{Xu-IEEE-T-CSVT}. It is also favourable if a single server can process video streams from various cameras. In a video surveillance systems, data are most likely compressed. Therefore, there is a direction that extracts features from compressed domain, such as \cite{Su2007MotionFlow,JingxinPRL,Xu-IEEE-T-CSVT,LiuVCIR2011,Manu2014,wang2008modeling}.

Though there are many different video compression standards available nowadays, the most widely used standard in surveillance applications are H.264, MPEG-4 Visual, and MPEG-2 \cite{Kruegle2011}. In addition, as it has been pointed out in \cite{Dey2013}, for network surveillance using H.264, the most common profile is the Baseline profile. The MPEG-2, MPEG-4 Visual and H.264 Baseline share similar frame structures. That is, they are all consist of three different types of frames, the I-frames, P-Frames and B-frames, where the I-frames contains appearance feature and the P and B frames contain motion vectors. Therefore, algorithms that are designed in one format is easy to be adapted to another. The most attractive characteristic for extracting features from compressed domain is the availability of motion feature in the compressed domain. This leads to a lot of efficiency as motion estimation is skipped compared to processing in image domain. The difference between motion vectors in compressed domain and optical flow is reviewed in \cite{Xu-IEEE-T-CSVT}, though there is still lack of literature to empirically evaluate the differences between the two. The main difficulty for such an evaluation is due to the variations of motion vector computation methods in video compression. Even for the same format, the motion vector computation is designed by the vendor of the cameras. Overall speaking, the motion vector estimation in video compression often share almost the same procedure of optical flow estimation in image processing, except they are computed in a lower resolution (the macro-block level). The process is faster as the encoding is done by hardware. Furthermore, in a surveillance systems, data are required to compressed for transmission and storage. That is, even an algorithm does not depend on features from compressed domain, the data are still required to compressed. The approach of extracting features from compressed domain can reduce the cost of video decoding as well. It is often helpful to save memory space. For example, in \cite{Manu2014}, it is shown that using compressed feature for action recognition can achieve a speed approximately to 100 times faster than  its counterparts in image domain. For fast processing of video, it is also possible to use GPU acceleration. However, as it has been identified in  \cite{Gregg2011} that there will be a bottle neck of PCI bandage for GPU programming on memory consuming applications.

The basic limitation of this approach is that, the algorithm is highly dependent on a set of similar compression standards, known as H.26x, and MPEG-x. Though these compression standards are popular nowadays, it is hard to expect whether they will still be popular in future. Once the compression standards have revolutionary changes, the algorithms may not be easy to update to catch up with the fashion. It is of interest to point out that the latest development of video coding standard  \cite{Gao2014} has begun to consider to incorporate features which can facilitate the development of intelligent surveillance. However, so far, those proposals have not been industrialised.

\subsubsection{Features Beyond Vision}

The discussions in the above sections solely focus on visual features. Besides visual information, other signals such as acoustic features can also be used.

Acoustic features have been popular in various surveillance applications, such as object tracking \cite{Beal2003,Shivappa2008} and event detection \cite{Transfeld2014,Moon2014,Valenzise2007,Foggia2015}. There are several benefits to use acoustic features. First of all, acoustic features are effective to detect events with significant audio characteristic such as a gunshot event. Second, visual information is easy to be occluded, whereas acoustic information is available even in the situation of serious occlusions. Foggia et al. \cite{Foggia2015} also suggested that audio surveillance systems can be deployed in locations where cameras are ``not allowed (public toilet)'' or in ``poorly illuminated areas''.

In \cite{Lefter2012}, a novel approach is proposed, which utilises both video and audio features to detect abnormal events in trains, such as ``harassment, hooligans, theft, begging, football supporters, medical emergency, travelling without ticket, irritation, passing through a crowd of people, rude behavior towards a mother with baby, invading personal space, entering the train with a ladder while the conductor is against, mobile phone harassment, lost wallet, fight for using the public phone, mocking a disoriented foreign traveller and irritated people waiting at the counter or toilet''. However, the dataset they used is captured based on a group of people acting the abnormal events. There is still lack of evidence for the level of success in the real situations.  A case study of using visual-audio sensor network for complex event detection is presented in \cite{Machot2011}. The algorithm presented in \cite{Machot2011} is a rule based detection algorithm for a``complex event'', which is an event that combines with basic visual actions and sounds. However, the details of detecting sounds and visual actions are not covered in the paper.  Some key information such as the number of instances of target events in the dataset  are missing. Therefore, evidence for the performance is not enough.

In a recent publication, Martin et. al \cite{Martin2015} proposed the a novel model called ``Weighted Finite-State Transducers (WFSTs)'' for the detection of acoustic events such as gunshot and glass-break in surveillance environments. The WFSTs model is in fact a combination of a set of Hidden Markov Models (HMMs), with each HMM for one type of events. The features tested in this reference are Mel-Frequency Cepstral Coefficients (MFCC), and its extensions such as Frequency Bank Coefficients (FBANK) and Mel-Spectral Coefficients (MELSPEC). However, such a system in \cite{Martin2015} is only able to detect the abnormal acoustic events, but cannot identify the physical location of the events. 

In a recent publication \cite{Foggia2015}, Foggia et al. proposed using acoustic features to detect traffic accidents such as car crashing in traffic surveillance. In this approach, audio streams are first divided into a set of partially overlapped short-time frames. Within each frame, acoustic features such as volumes, energy, zero crossing rate, and various spectral characteristics are extracted. Then the K-Means algorithm is used to cluster the high dimension features into a set of discrete words, based on which the histogram of words are computed (``bag of words''). Finally, support vector machine is used to classify the different traffic events. In addition, this article also presented an approach to set up microphones (i.e. determining the distance between two successive microphones).

A more popular approach for acoustic event detection is based on a set of microphone array. This is because microphone array is not only able to be used to detect detect an abnormal acoustic event such as a gunshot, but also able to identify the location of that event \cite{Valenzise2007,Moon2014}, which we termed ``acoustic localisation''. The principal for acoustic event localisation is related to some physics knowledge. The sound speed in a certain environment can be viewed as unique. Suppose the sound speed in the air at a certain area is \(v\).  There are microphones in the area at different locations, which form a microphone array. Suppose an acoustic event (e.g. gunshot) happens at a certain location, and generate a sound which transmit to the microphones. The microphone that is nearest to the event location will capture the sound at an earliest time, whereas the microphone that is farest away from the event will capture the information later than the others. The distances between the microphones are known, and based on this, an equation array can be set up to solve the location of the event.   

Besides video and audio, the signals that are generated by human interacting with machines are also useful in surveillance event detection. In \cite{dasani2015monitoring}, Dasani et. al propose a novel approach of applying process mining techniques, which originates from Business Process Management for abnormal operational event detection. In such an approach, a process model is built based on the standard operating procedure, and the operational event log in the infrastructure is analysed using some process mining tools to conduct a process conformance checking, so that the system can continuously monitor the actors' operations and detect abnormal behaviours.

It is interesting to mention the use of depth camera in video surveillance. The principal of depth camera is similar to radar detection. There are normally more than one normal cameras and an electromagnetic wave generator in a depth camera. The two cameras measures the time duration of receiving electromagnetic waves that are generated by the camera to estimate how far away the obstacle is.Therefore, using depth camera, one can get a traditionally RGB images and a depth image simultaneously. The additional depth dimension provides much convenience to interpret the motion in a three dimensional physic space. Recently, Bian et. al \cite{bian2015fall} proposed to use depth camera for fall detection. The silhouette based feature are used to model the fall event and normal events. With additional depth information, the silhouette reflects the physical reality in a more feasible approach, and thus improve the performance. In addition, as the electromagnetic wave emitted by the camera is infra-red light, the use of this type of depth images can detect the event when the light condition is dark. Besides the desirable characteristic at night time, in \cite{Bahnsen2015}, a fusion of information obtained from RGB and thermal cameras is shown to be used to detect road user actions, and the thermal camera is applied as it is robust to various weather conditions.

It is also possible to obtain depth information without an electromagnetic wave generator. For example, in \cite{Patino2015}, some stereo matching algorithm is used to get the depth information from three visual cameras (i.e. a trinocular camera). Through this trinocular camera system, Patino et al. proposed a method for detecting the abnormal activities (e.g. loitering) in waiting queues.

Finally, it is necessary to introduce the recently arising techniques that used Wi-Fi Signals for device-free activity recognition \cite{Zeng2015-ax,Pu2013-sr} The
term ``device free'' indicates that the persons do not need to wear any sensors. As it has been known that, the electromagnetic waves will have reflections, refractions and diffractions while transmitting through the atmosphere. Different locations and movements of human bodies will result in different fading characteristics of the wireless signals. Based on the channel state information that the received antenna received, it is feasible to extract features to recognize the activities. There are two main benefits of these approaches: 1) the radio signals can go through the visual obstacles and even go through a wall due to diffractions; 2) the wireless signals does not contain biometric information and thus preserve privacy.

\subsection{Classifying Unusual Events}
\label{sec:classifier}

Once feature vectors have been extracted to represent the events, those feature vectors are used as input to a classifier for detection. State-of-the-art pattern classification is heavily reliant on machine learning. In this section, we separate the discussion into the different machine learning paradigms. In Section \ref{unsupervised} we review methods based on unsupervised learning; in Section \ref{supervised} we discuss the methods on supervised learning; finally, we discuss weakly supervised methods in Section \ref{weakly supervised}.


\subsubsection{Unsupervised Learning Approaches}\label{unsupervised}

The most widely used paradigm is unsupervised learning \cite{Adam2008,Andrade2006,cui2011abnormal,GaoYang2011,li2011learning,Mehran2009,Wang2009PAMI,wu2010chaotic,yang2011,Zen2011}. The unsupervised learning approach in this field is typically an anomaly detection system, where the unusual events are defined as those with a low probability in the training datasets. Many models that can be used for anomaly detection have been applied in the literature,
including Gaussian Mixture Models (GMMs) \cite{Ryan2011}, Hidden Markov Models (HMMs) \cite{Andrade2006,Kratz2009,XiangG06}, Neural Networks \cite{yang2011}, and Probabilistic Topic Models (PTMs) \cite{hospedales2009markov,li2011learning,Wang2007,Wang2009PAMI}. 

There are several practical reasons for the popularity of this strategy. Firstly, the term ``unusual events'' implies the low probability of these events, and the events of security interest often occur with lower frequency compared to others. Second, event annotation on large scale surveillance video is impractical, especially the marking of bounding boxes to annotate where the event occurs. 

However, the suitability of these models depends on the distribution of the data, and also relates to the feature that has been used to represent the events.

The Gaussian Mixture Model assumes the data is drawn from a mixture of Gaussian distributions. Once it is trained, it will be very efficient in computing the probability of a sample input from the test data. The low probability patterns are detected as the unusual events \cite{Ryan2011,wu2010chaotic}. Besides the efficiency in the detection step, a significant benefit of this model includes the capability to support high dimensional continuous inputs, especially when there are dependencies among the channels of feature inputs as these dependencies are captured in the covariance matrix. However, due to the large number of parameters in the covariance matrices, this model can be prone to over-fitting. When there is insufficient data to correctly learn all parameters, the model becomes tuned to the training data. Especially when the data distribution is non-Gaussian, in order to learn GMMs, we need to set the number of mixtures to a large number \footnote{Typically, a GMM can be used to model any kind of distributions. However, some distributions are hard to model using a limited number of mixtures. Setting the number of clusters into a large value leads more severe over-fitting.}. Furthermore, when there is a cluster lacking enough samples, the covariance matrix may become singular, leading to numerical errors. In order to overcome this issue, one can perform a dimension reduction in the pre-processing stage. Alternatively, one can use a diagonal covariance matrix. However, either approach will cause a loss of information. For these reasons, GMMs are limited to applications where the feature vectors are low dimensional, relative to the number of training samples. 

GMMs are sometimes also used together with Hidden Markov Models. Andrade et. al \cite{Andrade2006} proposed an algorithm using a Mixture of Gaussian Hidden Markov Models (MOGHMM) to detect anomalies in crowd scenes. The pixel intensity is viewed as a random variable. It falls into the Mixture of Gaussians (MOG) distribution. Instances with very low probabilities are deemed to be abnormal. However, there are no explicit meaningful hidden states in their application. Since the MOG-HMM is a combination of a GMM and HMM, if the GMM can detect abnormal events, this model should have the capabilities as well. However, there is lack of evidence that the more complex MOG-HMM works better. Given that it is hard to collect a dataset with a significant number of unusual events in a controllable manner, the evaluation of this algorithm depends on simulated videos generated by computer graphics.  Nallaivarothayan et. al. \cite{Nallaivarothayan2013} proposed using a 2-D semi-HMM to detect the outliers in video surveillance, and features such as optical flow, location and flows are used. The method is shown to be effective to detect local abnormalities in the UCSD dataset. 

Xiang and Gong \cite{XiangG06} present a unified bottom-up and top-down automatic model selection based approach to model activities beyond tracking. In this algorithm, object-independent events are segmented using automatic model selection based on Schwarz's Bayesian Information Criterion (BIC) (bottom up). The Dynamically Multi-Linked Hidden Markov Model (DML-HMM) for behaviour understanding is built using BIC-based factorization resulting in its topology being intrinsically determined by the underlying causality and temporal order among events (top down). For measuring multi-scale temporal changes at each pixel, pixel change histories based on an accumulative and decay model are used. This is one of the earliest works which clearly identifies the significance of modelling activities beyond tracking. Though in \cite{XiangG06} the need of methods beyond tracking for crowded scenes has been well-addressed, however, the dataset used for evaluation in this paper is rather sparse (only several people and objects) compared to many real world crowded scenes (hundreds of people).

Utasi et. al \cite{utasi2010detection} presents a framework to detect unusual flow patterns using multi-level Hidden Markov Models. A histogram of the directions of optical flow is built. Mean-shift segmentation \cite{georgescu2003mean} is applied to segment the image into different regions of interest based on the histogram of directions. Meanwhile, foreground subtraction is performed by a robust adaptive background subtraction algorithm \cite{stauffer2000learning}. The moving pixels form connected components in the scene. Inside every region of interest (connected component), image segmentation is performed using a Gaussian Mixture Model (GMM) for a lower level segmentation of each connected component, using the EM algorithm. The mean vectors of GMMs are applied to be the input to HMM. The segmentation algorithm is performed on each connected component instead of the whole region of interest, which avoids the loss of information around small objects. The algorithm assigns an HMM to each region to detect anomalies. The emission probability represents the likelihood of a pattern in the test data, and it is used as the detection criterion. The hidden states in this application are driven by traffic lights contained within the data, giving them a physical meaning which is lacking from other approaches such as \cite{Andrade2006}.
However, the unusual events in the test video are manually edited, making them lack of the real-world nature.

HMMs are usually used to model the temporal correlations of the activities, but this is not always the case. In \cite{Kratz2009}, features from spatio-temporal patches are extracted and HMMs are used to model the sequential dependencies among patches. Besides temporal correlations, the spatial correlations are modelled as well. Though HMMs have demonstrated a level of success in modelling activities with correlations in time and space, the correlation has to be in one direction. In graphical model theory, HMMs are directed graphs (Bayesian Networks) \footnote{There are fundamentally two classes of graphical models, the Bayes networks and Markov Models. The difference between them is that the Bayes networks are directed graphs and Markov Models are undirected graphs. HMMs are Bayes Networks since they are directed graphs and can be inferenced using Bayes rule directly. Readers should not be confused by this name. Typical Markov Models in Graphical Models are Markov Random Fields}. That is, there is only one direction in the state transition. This limits HMMs to modeling more complicated events with mutual correlations in space. HMMs are more successful in modelling temporal dependencies for the events as naturally there is a causality property for the events in surveillance scenes. However, it is not rare to have complicated spatial correlations for the events in the same scene. Thus the suitability of using HMMs to model spatial correlation is questionable. A potential solution for this problem is the use of Markov Random Fields (MRF) for unusual event detection \cite{Kim2009}. In \cite{Kim2009}, Principal Component Analysis (PCA) is used to model local events and these local events are modelled by MRFs for a global analysis. Though this approach is shown to be able to detect some complicated events, there is still a significant missed detection rate. Furthermore, a lot of events with no security interest (i.e. cleaning the floors) are identified as unusual events as well. 

Recently, probabilistic topic models \cite{Blei2010} have gained widespread use in the applications of activity perception after the pioneering work of \cite{Wang2009PAMI}. In \cite{Wang2009PAMI}, it is identified that the benefits of the topic model are the capability to learn co-occurring events (notebly the jay-walking events in traffic surveillance). However, the ability to model coexisting events is not restricted to topic models provided suitable feature representations that capture interactions are used. Thus we argue that the benefit of topic models is the capability of modelling multi-agent interactions with simple feature descriptors, which do not inherently capture the interactions. 

\begin{figure}
\begin{centering}
\includegraphics[scale=0.6]{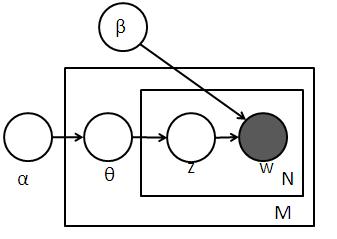}
\par\end{centering}

\caption[Latent Dirichlet Allocation]{Latent Dirichlet Allocation \cite{Blei2003} \label{LDA}}

\end{figure}

Figure \ref{LDA} provides a graphical illustration of Latent Dirichlet Allocation \cite{Blei2003}, which is a generative probabilistic topic model originally proposed in natural language processing. A corpus is formed by $\mathrm{M}$ documents; each document is a bag of words; the words are generated by $\mathrm{K}$ topics from a fixed vocabulary. Hence, each topic corresponds to a distribution over the vocabulary. The word ($\mathrm{w}$) is the only observed variable in this model. $\beta$ is a matrix which stores the multinomial distributions of words from the vocabulary for the $\mathrm{K}$ topics, and $z$ is the label of the topic for each word. $\mathrm{\alpha}$ is the Dirichlet parameter and $\mathrm{\theta}$ is the topic distribution for each document, which draws from the Dirichlet distribution,

\begin{equation}
\mathrm{p(\theta|\alpha)=\frac{\Gamma(\sum_{i=1}^{k}\alpha_{i})}{\prod_{i=1}^{k}\Gamma(\alpha_{i})}\theta_{1}^{\alpha_{i-1}}\cdots\theta_{k}^{\alpha_{k-1}}.}\label{eq:Dirichlet Distribution}\end{equation}

Latent Dirichlet Allocation (LDA) is a finite mixture model, where the number of topics must be known beforehand. This limitation can be overcome by using Hierarchical Dirichlet Processes (HDP) \cite{Teh2006}, which is a non-parametric Bayesian model. In HDP, the number of topics is assumed to be infinite, but within a finite size corpus, there is a finite number topics.

In applying probabilistic topic models such as LDA and HDP to activity perception \cite{Hospedales2011,li2011learning,Wang2007,Wang2009PAMI}, the most popular feature descriptor is that the location and direction information (computed by optical flow) of each moving pixel are decoded into a discrete word. A long video sequence is viewed as the {}``corpus\textquotedblright{}, and is divided into a series of small video clips termed {}``documents\textquotedblright{}. The activities are the {}``topics\textquotedblright{}. The whole video sequence shares the same $\mathrm{K}$ activities, but the distribution of the $\mathrm{K}$ activities is temporally different. As hierarchical Bayesian models, LDA and HDP model two levels of activities: the distribution of visual words, and the distribution of activities (topics). As a result, through detecting the abnormal pixel motion patterns, they detect the local anomalous events; and by computing the activity distribution, they detect abnormal video clips and video clips with unusual co-existing events. Because this approach does not rely on object tracking, it is applicable in crowded and complex scenes. 
The fundamental assumption for topic models is the ex-changeability of words over a document. The property of ex-changeability is often called the ``bag of words'' assumption, and implies a loss of temporal information as the order of the words is not considered.

The limitations of \cite{Hospedales2011,li2011learning,Wang2007,Wang2009PAMI} arise from two aspects:
\begin{enumerate}
\item ignoring the correlations of words in a document may not fit reality; and
\item there is a gap in matching the concepts such as ``corpus'', ``word'', and ``documents'' to the field of computer vision. The partition of video into clips means an event may be split over multiple clips. The quantization error in the process of converting high dimensional continuous features into discrete values also causes potential inaccuracies. The majority of topic models only supports discrete inputs.
\end{enumerate}

There have been approaches \cite{hospedales2009markov,kuettel2010s} to address the first issue, by exploring the techniques from HMMs. In \cite{kuettel2010s}, HMMs are used after the learning of atomic activities in an extension of HDP named ``Dependent Dirichlet Processes''. In \cite{hospedales2009markov}, the first Markov assumption is used to model the dependencies between two successive video clips and an extension of LDA named ``Markov Clustering Topic Model'' is proposed.  It is shown in \cite{hospedales2009markov} that the new model outputs either HMM and LDA in detecting abnormal events using the QMUL traffic dataset. However, there are few approach to addresses the second problem listed above.

The learning approaches reviewed in this section so far are all graphical models, where the GMMS, HMMs, and topic models are Bayesian networks; and the Markov Random Field belongs to Markov models. Though graphical models have achieved a great popularity in this field, there are still many other methods which are active in the research including KNNs (K Nearest Neighbours) \cite{Tziakos2010}, SVMs (Support Vector Machines) \cite{Tziakos2010} and neural networks \cite{yang2011}.

In \cite{Tziakos2010}, both KNNs and one class SVMs are used in the application of abnormal event detection in a subway dataset. The KNN method works slightly better than the one class SVM. Though the method can be used to detect some abnormal events efficiently, the occurrence of unusual events in the evaluation does omit the event of passenger passing the entrance of the subway without payment, which is in fact the event with largest security interest. 

In \cite{yang2011}, the features are extracted using a spatio-temporal framework. The histogram of optical flow in a space-time grid is used as the feature descriptor. The classifier for event detection is an online sparse coding proposed in \cite{yang2011}. In this work, a motion pattern is represented as a linear combination of a sparse coefficient vector over a set of basis functions. The sparsity of the coefficient vector means most entries are $0$. The learning process trains the basis functions on normal behaviours. Given an observation and the learnt basis functions, a sparse approximation method is used to compute the coefficient vector and the reconstruction cost. Because the basis function is trained on normal events, for the motion pattern of unusual events, the reconstruction cost is expected to be high. Based on a threshold of the reconstruction cost, the unusual events are detected. Similar approaches also appear in \cite{Zhu20141791,Xu2011}. In \cite{Xu2011}, the local binary patterns from three orthogonal planes (LBP-TOP) is applied as the feature representation, which captures more appearance features. The importance of whitening as a preprocessing step is also highlighted in \cite{Xu2011}, due to the fact that the Euclidean distance is normally used in sparse coding.  More recently, in \cite{Zhu20141791}, the Earth Mover distance is employed and shown to have better performance compared to the state of the art.


Typically, novelty detection techniques are used in the situations where the usual events are simple in form, and the number of unusual events is large. Only when the usual events are simple in form, can a model that represents the universe of all usual events be trained. For instance, one can detect all the behaviours that are not ``pedestrian walking''. In this example, the usual events are only pedestrian walking, while the unusual events can be person running, person jumping, animal running, vehicle moving and person skating. 

\subsubsection{Supervised Learning Approaches}\label{supervised}

The unsupervised learning approach typically is used in the case when the events we want to detect are low probability events. However, events with security interest are not always rare. In addition, we may want to find a certain kind of event with prior knowledge. For instance, the events which break traffic rules have been defined and are known. In such cases, we can consider adding a level of supervision to the learning process \cite{Kellokumpu08humanactivity,LiuVCIR2011,Yuan2009,2011TokyoCanon,Takahashi2010,2011PKU}. 

The supervised learning approach is less popular compared to its unsupervised counterpart. The main problem is the difficulty in labelling the target events in the training dataset. For action recognition in non-crowded scenes with clear backgrounds, a temporal level annotation is sufficient as one frame only contains one event \cite{Kellokumpu08humanactivity}. In crowded scenes, the annotation becomes challenging as there are many events co-occurring and a precise annotation requires the marking of bounding boxes.

The strategy of a sliding window is used widely within the supervised learning framework \cite{Yuan2009}. A target model built during the training process. Then this model is compared to a sliding window for the detection of the target pattern. Because the window size is relatively small compared to the whole image, this strategy allows the matching process to operate over a relatively small region with only one event. Though in \cite{Yuan2009} the problem of computational trade-off has been solved by a novel sub-volume search algorithm, there are still limitations such as the failure to detect interaction level events. Meanwhile, the window size is difficult to set and, depending on the scene, it may not be appropriate for it to be fixed due to perspective distortion. To train such a target model, one can use a dataset with a clear background to learn the action, avoiding the need to mark bounding boxes. However, as the camera views and surveillance scene change, detection becomes more challenging as the training data does not match the scene in the test dataset.

Therefore, there is a concept called ``cross-dataset event detection'' proposed in recent years \cite{Cao2010}, which aims to detect events in the test dataset with different scenaries of the training data. In \cite{Cao2010}, Gaussion Mixture Model is used. A universal background model (GMM) is trained on a large training dataset. There will be a small dataset with the same scene of the test dataset used for adapt the GMM to the specified scene of the test data. Then the final adapted model can be used to detect events.  However, this approach typically require a quite large dataset to train the universal background model. Similar techniques have been used in speaker verification for a long history.

In \cite{Ji2010}, a multiple layer neural network is proposed. The algorithm starts by detecting the heads of pedestrians in a crowded scene, and then by tracking the heads, raw pixels from the 3D volumes around the heads are used as the input features to a multilayer neural network. The early layers in this model are viewed as feature extraction. The top layer of this model is the classifier. In the literature, such a model (multi-layer neural network with raw pixels as the input) is a deep learning architecture (unsupervised feature learning).  However, this method relies on robust head detection in crowded scenes. A missed detection of a pedestrian's head will result in a missed detection of an event. In surveillance event detection, a missed detection is more harmful than a false alarm. In addition, this method requires special computational resources (i.e. parallel computing, GPU, multi-cores system, multi-threaded programming) and even with such resources, the computational time is very high. For this reason, this is not a practical method for real world applications at the present time.

More recently, Shao et al. \cite{Shao2016-qr} proposed a so called ``slicing
3D convolution neural network'' for crowd analysis, where convolution neural network is
applied in three different filters of the video volume (based on ``xy'', ``xt'' and ``yt'' coordinates). Then the feature vectors learned from the three CNNs are concatenated into a single feature vector, which is then set as the input for a linear support vector machine. The parameters are initialized as a pre-trained model that learned from the ImageNet dataset. It is quite intuitive that the pre-trained parameters for ImageNet should be suitable for the CNN for the ``xy'' plane, as the images from ImageNet are natural images with similar characteristics to the images from the ``xy'' plane. However, the nature of the image data from the other two planes is different from that from the ``xy'' plane. Nevertheless, Shao et al. \cite{Shao2016-qr} has showed that this approach achieved superior performance for crowd attribute prediction rather than event detection. Compared to \cite{Ji2010}, the feature in \cite{Shao2016-qr} is proposed in a more unsupervised manner, as the parameters are not initialised through hand-crafted features, and the deep learning model model is not applied on top of a tracking by detection framework.  

Some very complicated events are defined as a set of atomic actions following a certain protocol. For such events, it is possible to define a graph to model the structure, and then techniques are developed to detect events that violate this structure. A popular tool to model these rule-based protocols is Petri net. It is especially useful to handle complicated events with a lot of sequential and concurrent dependencies. References using Petri net for surveillance event detection include \cite{Albanese2008,Borzin2007,Ghanem2004,Lavee2013}.  Recenly, Cheng et al. \cite{Cheng2014} proposed to use a model termed a ``sequence memoizer'' to learn the temporal dependencies and identify the events. For instance, a splitting event is typically followed by a meeting events in an airport, as a group of people have to first meet together before being able to split up. However, algorithms relying on such assumptions may be vulnerable in the real world, as abnormal event typically does not follow the normal path.

In summary, supervised learning approach is useful in the situation where we have a specified event to detect. It is vulnerable in situations where the unusual events are unknown.

\subsection{Weakly Supervised Learning Approaches}\label{weakly supervised}

The unsupervised learning approach often generates a lot of false alarms especially if the training dataset is small, as any events that are not in the training dataset will be identified as the abnormal events but they may be not; on the other hand, the supervised learning approach often encounters difficulties in the labelling of the training dataset. This section reviews the methods that use a third alternative, that is, weakly supervised learning.

The phrase of ``weakly supervised learning'' has been appearing frequently in the literature of machine learning to indicate a methodology between the supervised learning and the unsupervised learning approaches. However, there is a lack of a widely accepted definition for the concept of weakly supervised learning. As a result, the concept of ``weakly supervised learning'' varies in different situations. Meanwhile, there are a set of similar terminologies such as ``semi-supervised learning'' and ``multiple instance learning''. In some situations, these terminologies are interchangeable and in other situations they are not. To facilitate the discussion, we use the phrase ``weakly supervised learning'' within the thesis based on a domain dependent definition which follows a recent publication \cite{Hospedales2011}: The data is partitioned into a set of video clips. A video clip can contain multiple events. There are binary labels to the video clips in the training dataset for a specified event, with label ``1'' indicates the presence of the event, and label ``0'' indicates the absence of the event. However, the annotation does not contain fine information for when and where the event happens and what the event is. A classifier is trained based on the binary labels at the clip level. Given a video clip from the test dataset, the classifier is required to predict the presence or absence of the event of interest.

In non-crowded scenes, each video clip may contain only a single event, and thus there are no ambiguities. However, in the situations of crowded scenes, a video clip typically contains a lot of events. Among them, it is possible that only one to two events refer to the event of interest. The number of background events is usually much larger than the number of events of interest, whereas the background events in the video clips labelled ``1'' are typically considered as the noise to the useful signal (event of interest). Traditional supervised learning techniques are often developed without any consideration of the presence of noise in the sample input. Thus the theories are typically constructed under an implicit assumption that the sample input does not contain noise. This leads to the failure of most supervised learning approaches when only ``coarse level labels'' are available for event detection in crowded environments and justifies the need of weakly supervised learning approaches in these situations. 

Hospedales et al. \cite{Hospedales2011} proposed a weakly supervised joint topic model for rare event detection (vehicle turns in traffic footage). In \cite{Hospedales2011}, the video is divided into uniform clips and there are binary labels at the clip level to indicate the presence of an event. A topic is viewed as a distribution over the code words by encoding the location and motion direction of the moving pixels. However, the initialization of parameters for this model is difficult and the features to represent the events are based on optical flow, which is poorly suited to capturing long duration motion information.  It is shown in  \cite{Hospedales2011}
 that the weakly supervised joint topic model outperforms its unsupervised learning conterpart (LDA) and its supervised counterpart (SVM) in the detection of subtle behaviour in traffic surveillance. 

Xu et. al. \cite{Xu-IEEE-T-CSVT} extends the Labeled LDA from language processing into the application of surveillance event detection. Compared to the WSJTM, labeled LDA has fewer parameters to train, and potentially overcome the overfitting issue of the WSJTM for small dataset, but if the dataset is large enough, the labeled LDA approach is easier to have underfitting. In  \cite{Xu-IEEE-T-CSVT}, experiments shows promising performance of detecting the People Meet, People SplitUp and Embrace events in the TRECVid SED dataset.

In \cite{JingxinPRL}, an algorithm is proposed, which uses random sensing and orthogonal matching pursuit (OMP) to support MIL for event detection, which is tolerant to a limited number of positive training samples but requires sufficient negative training samples. This approach is called ``MIL compressive sensing''. Here the term MIL is exchangeable to ``weakly supervised learning''. This method is novel and has shown to be able to achieve well performance in the detection of traffic turns events. Though its performance for Jay Walking detection is limited, it still outperforms many others. However, the theory is constructed by some assumptions which may be too strong in real world applications. One requirement is that the video clips labelled ``0'' should contain
all the background activities. In practice, this requires sufficient training samples
labelled ``0''.  
This method also requires the event of interest to be independent
from the background activities. However, it may be hard to determine
if this criteria is met in some situations. If we know that the event
of interest is able to be described as a distribution over a disjoint
subset of the vocabulary, it is likely that it can be separated from
other events independently.
 However, such restriction is strong in the real world and limits it to be a widely accepted approach.
 
\subsubsection{Active learning}

The terminology ``active learning'' means that, a learning algorithm will request the users to provide feedbacks for the learning results and based on which, update the algorithm. From this view point, it can be categorized as a special type of online learning.

In \cite{Hasan2015}, an approach of using active learning for activity recognition is proposed. The procedure is that, the algorithm is first initialised with a small set of labelled training samples, which is much smaller than the overall training dataset. Then the algorithm will update the labels of other unlabelled training samples. The algorithm will automatically assign labels to some samples with a confident detection result, whereas it will request the user to label other samples. The labels marked by human are termed as the ``strong teacher'', whereas the labels marked by the algorithm are termed as the ``weak teacher''. Since most labels are marked by the algorithm, this approach can relieve the burden of human annotation to some extent. However, this approach requires human interacting with the system, making the application more challenging to use. In addition, the algorithm has to be run in real time even in the training process as it interacts with people, which restricts the freedom of available techniques. It is also hard to determine which sample should be labelled by human. These difficulties decrease the performance of the overall algorithm. A further study of this approach is presented in \cite{Hasan2015}.

\subsubsection{Summary}

This section summarizes the machine learning methodologies for surveillance event detection. In terms of the level of supervision, we separate the discussion into three categories: unsupervised learning, supervised learning, and weakly supervised learning.

Unsupervised learning is most widely used in situations where we don't know the target abnormal events, and the target events may be in a large number. 
Supervised learning is useful in situations where we have a wide range of normal behaviours and only a few abnormal events in sparse background. Weakly supervised learning approach is a special type of supervised learning approach, but it is more suitable in crowded environments as it requires only coarse labels. Table \ref{tab:machine learning} illustrates about how to select the suitable machine learning approaches. 

\begin{table}[H]
\caption{Comparison of different machine learning methodologies\label{tab:machine learning}}{
\begin{tabular}{|>{\centering}p{3cm}|>{\centering}p{2cm}|>{\centering}p{2cm}|>{\centering}p{3cm}|>{\centering}p{3cm}|}
\hline 
Type & Num. Target Events & Num. Regular Events & Annotation in Training data & Level of Crowdness\tabularnewline
\hline 
\hline 
Supervised Learning & few & many & fine annotation  & sparse and crowd (single local or global event)\tabularnewline
\hline 
Unsupervised Learning  & many & few & no annotation & sparse and crowd\tabularnewline
\hline 
Weakly Supervised Learning & few & many & coarse annotation & crowd (target events and regular events co-occurring)\tabularnewline
\hline 
\end{tabular}
}
\end{table}

\section{System Integration as a Product}
\label{sec:system}

Surveillance event detection is mostly a research intensive topic with many problems unsolved until now, whereas surveillance application has been widely used in many aspects in our lives. For a complete review of this topic, it is necessary to review how the algorithms proposed in research are integrated into a real world product. In large facilities such as transportation hubs, the surveillance systems are built on top of workflow management systems \cite{Szwed2013}. In \cite{Machot2011}, a framework of integrating surveillance event detection into a workflow system is presented. In \cite{Szwed2013}, the architectures of several industrial surveillance solutions are presented. A workflow management system that aims at surveillance applications is called ``physical security information management''. There are several industrial standards for surveillance related products, such as the ``Open Network Video Interface Forum (ONVIF)'' \footnote{http://www.onvif.org/} and the ``Physical Security Interoperability Alliance (PSIA)'' \footnote{http://www.psialliance.org/org.html}.

Below lists a few real world surveillance products with automatic event detection functionalities:
\begin{itemize}
\item Bosch IVA (Intelligent Video Analysis) \cite{BoschIVA}: This system allows the user to specify the rules for the target events, such as location, shape, object size, etc, and then the system can automatically alarms the target events if they occur. Example target events include line crossing, loitering, removing objects, and so on.
\item D-ViewCam Plus (Advanced surveillance management Software) \cite{D-viewCam}: This product manages alarms from different types of sensors including cameras, smoke detectors, emergency buttons, card readers and so on. Optionally, the users can install the intelligent video analytics packages, which can be used to detect an object being inside or crossing a zone.
\item Axis Intelligent Video \cite{Axis} The Axis intelligent video software system include functionalities of detecting people manipulating the cameras, audio detection based on the volume, and video motion detection. 
\item Aimetics Video Analysis \cite{Aimetis}: This software platform provides functionalities such as wrong direction detection, abandon object detection, loitering and etc. 
\end{itemize}
From the above discussion, we can see that there has been industrial products with event detection functions in surveillance systems. However, the events that can be detected are all very simple ones that can be determined using a simple rule on a simple feature (e.g. moving direction, audio volume, motion detection. etc.). It remains an empty area in industrial products for complicated activity recognition. 

\section{Discussion}
\label{sec:Discussion}

A system for unusual event detection consists of two fundamental parts: feature extraction for event representations, and classifier design for event detection. 

The purpose of feature extraction is to find a compact representation of the events which is able to separate the unusual events from usual events. In non-crowded scenes, the feature vectors are often designed based on objects' trajectories. However, due to occlusions, object tracking often fails in such conditions. As a result, most state-of-the-art methods in this topic adopt local features to represent the events. Though local features are robust in crowded environments, they do not model events that span large areas of the scene, or occur with a long duration. The particle trajectories capture motion information with long duration and large space while they are more robust when compared to objects' trajectories. 

To classify unusual events, there are three main approaches: unsupervised learning; supervised learning; and weakly supervised learning. Unsupervised learning approaches are most widely used, simply because of the difficulty in listing all the unusual events beforehand in many surveillance applications. Typically, the number of possible unusual events is infinite. Meanwhile, the probability of an unusual event is often low. In the unsupervised learning approaches, a training dataset without any unusual events is prepared. A statistical model is trained for the usual events. The unusual events are detected as the low probability events based on the trained model. However, due to the diversity of the usual events, it is very hard to train a single statistical model which can model all kinds of usual events. Limited by the size of the training dataset, it is possible to have a usual event in the test dataset that is not observed in the training dataset. In such situations, there will be false alarms. A typical problem for supervised learning is the difficulty in annotating ground truth in the training dataset. A video frame can contain many activities. In a crowded environment, an unusual event often co-occurs with a set of background activities. To correctly label the unusual events and filter out the noise caused by co-occurring usual events, the tedious marking of bounding boxes is required, something which is impractical for any moderatley large dataset. Recently, a concept called ``weakly supervised learning'' has been introduced in the literature. The weakly supervised learning approach uses coarse level label (frame level or clip level) annotations to denote the presence of an unusual event. In the detection process, the trained model predicts whether a video clip or a frame contains the event of interest or not. 

This research topic is a highly undeveloped one \cite{Porikli2013}. The main difficulties are in data collection and experiment evaluations. Regarding the problem of data collection, it is not practical to collect a dataset in a real world environment with the unusual events with security or safety interest performed in a controllable manner. Due to this reason, the events identified in state-of-the-art datasets are not the unusual events in the common sense for surveillance purposes. There is an exception in \cite{Krausz2012}, where a case study applying optical flow based technique to analyze video footages from a real accident (the ``Love Parade 2010'' event) is presented. However, the dataset is not publicly accessible by researchers in the community. Meanwhile, the datasets in this field are often too small for a statistical result. Although many datasets contains several hours of footage or many thousands of frames, the number of unusual events contained in often very small. In the datasets that we review in this paper, only the TRECVid dataset can be viewed as one that can generate results with statistical significance. However,  even the TRECVid dataset has significant limitations. For example, the events identified  in the TRECVid datasets such as people meeting and people splitting are normal behaviours in any airport. There is a lack of events defined in the dataset that are of real security interest. The UMN dataset and PETS 2009 datasets are performed by actors and the events are not natural. The UCSD dataset is too small, and the events defined in this dataset are relatively simple. The MIT traffic dataset does not have official ground truth. %
Many publications using this dataset such as \cite{Wang2009PAMI}, do  not publish the ground truth they use. Other researchers \cite{Hospedales2011} annotate the dataset, but do not released it to the public. 
As such, it is not uncommon to have one dataset with several different ground truth annotations, and algorithms using this dataset are therefore hard to compare with each other. 

The evaluation protocol is another critical problem in this field. An event will span a set of continuous frames; however, it is rare for an algorithm to detect all the frames containing the event. In reality, if a system detects one frame and fires an alarm for an event, it can be viewed as performing as expected in the detection of this event. However, in simply counting how many of such frames are detected, there will be a lot of missed detections. So far, there is still lack of a widely accepted yet scientific evaluation methods.

So far, there is no methodology that can achieve significantly better performance than others in this field. Because of this, a wide range of diverse techniques have been proposed and active research is currently being conducted in this field. Nowadays the CCTV surveillance systems still almost solely rely on human operators to monitor the scenes and detect unusual events. 

In future, the problem of data collection and annotation may be solved by a combination of computer graphics and computer vision techniques.  With the development of computer graphics and computer simulation, there have been many excellent virtual reality platforms  such as the OpenSim platform \cite{OpenSim}. In the future, it would be  possible to first generate video with simulated  events and then control the occurrence of the events by software programs. In this way, we can have plenty of data with an event of interest inserted in  a controllable  manner. There  have been a few publications in the area of event detection using artificially  generated videos  \cite{Andrade2006,utasi2010detection}. There are also some novel idea about using virtual reality data for pedestrian detection \cite{Hattori2015}, which is normally an initial step for object tracking and event detection.   Another technique proposed for generating artificial data is to simulate crowd movement using the social force model. Object tracking is used to acquired real video  footage  to capture trajectories and these trajectories are used to train the parameters for a pedestrian behaviour model for simulation purpose  \cite{Lee2007}.  

A combination of computer vision and computer graphics techniques is expected to be used in the future to artificially generate very  large databases with large numbers of realistic unusual events. These databases, together with fully developed evaluation protocols, will enable effective algorithms to be developed to detect unusual events in crowded environments with a level of accuracy required for real world deployment. Currently, there has been some publication \cite{Simone2014} in this direction. An alternative direction is to use radio control toys to simulate abnormal events such as that presented in \cite{Hu2006}. The challenge in this direction is about how to make the toys' movements close to real movements of pedestrians and vehicles.  For such a direction, it also highly integrated into the development of control systems and robotics.

One problem of video surveillance research is the increasing attention of personal privacy. Therefore, recently there is also some article for surveillance data anoynimisation \cite{Birnstill2015}. In Birnstill et al. \cite{Birnstill2015}, the proposed approach applied image process techniques to remove personal identifiable information on videos while preserving the information for event detection. However, normally we also want to know who conduct the suspicious events. Therefore, it is still not an ideal approach for video surveillance with privacy protection. The use of Wi-Fi Signals \cite{Pu2013-sr} rather than visual information is a direction for privacy preservation. 

The present review also shows that besides visual sensors, microphones, thermal cameras, radar and Wi-Fi devices are significant to surveillance applications as well. However, the integration of different signals from various types of sensors is still an open problem until now, which can be a future research direction.

\section{Acknowledgement}
We would like to thank Dr Simon Denman for his recommendations to improve the paper.


\bibliographystyle{plain}
\bibliography{review_paper_ref}

\end{document}